\pdfoutput=1
\documentclass[conference]{IEEEtran}
\usepackage{cite}
\usepackage{amsmath,amssymb,amsfonts}
\usepackage{algorithmic}
\usepackage{graphicx}
\usepackage{textcomp}
\usepackage{xcolor}
\usepackage{booktabs}
\usepackage{multirow}
\usepackage{url}
\usepackage{placeins}
\def\BibTeX{{\rm B\kern-.05em{\sc i\kern-.025em b}\kern-.08em
    T\kern-.1667em\lower.7ex\hbox{E}\kern-.125emX}}

\begin{document}

\title{Optimizing H-Graph Hybridization for Diffusion-Guided RRT}

\author{\IEEEauthorblockN{Omer Talmi}
\IEEEauthorblockA{Tel Aviv University\\
omertalmi@mail.tau.ac.il}
}

\maketitle

\begin{abstract}
Sampling-based motion planners guided by diffusion models produce high-quality trajectories in a single run, yet the stochastic diversity available at inference time is left largely unexploited. We present two inference-time diversification strategies for a fixed, pretrained DiTree model, combined via H-Graph hybridization, and evaluate them on a holonomic AntMaze robot across 15 maze scenarios. The first, \emph{factorial} diversity, sweeps the random seed and Diffusion Goal Bias (DGB) parameter, the second, \emph{refinement-only} diversity, sweeps the diffusion refinement strength (RS) that controls how much an RRT-generated trajectory is edited. Because a single-run baseline only partially succeeds, we additionally compare H-Graph results with pool-based statistics. H-Graph improves the mean pool length of the factorial and refinement-only diversities by 18.8\% and 14.5\%, respectively. In addition, it also improves the best individual candidate's lengths by 9.7\% and 6.8\%, respectively. And last, compared with the successful baseline's trajectory length, it improves the results by 18.2\% and 19.9\%, respectively. These results show that inference-time parameter variation is a reliable, training-free source of path diversity, and that H-Graph hybridization reliably converts this diversity into shorter, higher-quality trajectories.
\end{abstract}

\begin{IEEEkeywords}
motion planning, path planning, trajectory optimization, diffusion models, mobile robots
\end{IEEEkeywords}

\section{Introduction}

Kinodynamic motion planning (KMP) \cite{lavalle2006planning} remains a fundamental challenge in robotics, requiring the generation of feasible trajectories that satisfy both kinematic and dynamic constraints while avoiding obstacles. Recent advances have shown promise in leveraging learned diffusion models toward action sampling in RRT-style planners \cite{hassidof2025trainonceplananywherekinodynamicmotion}.
\begin{figure}[!t]
\centering
\includegraphics[width=\columnwidth]{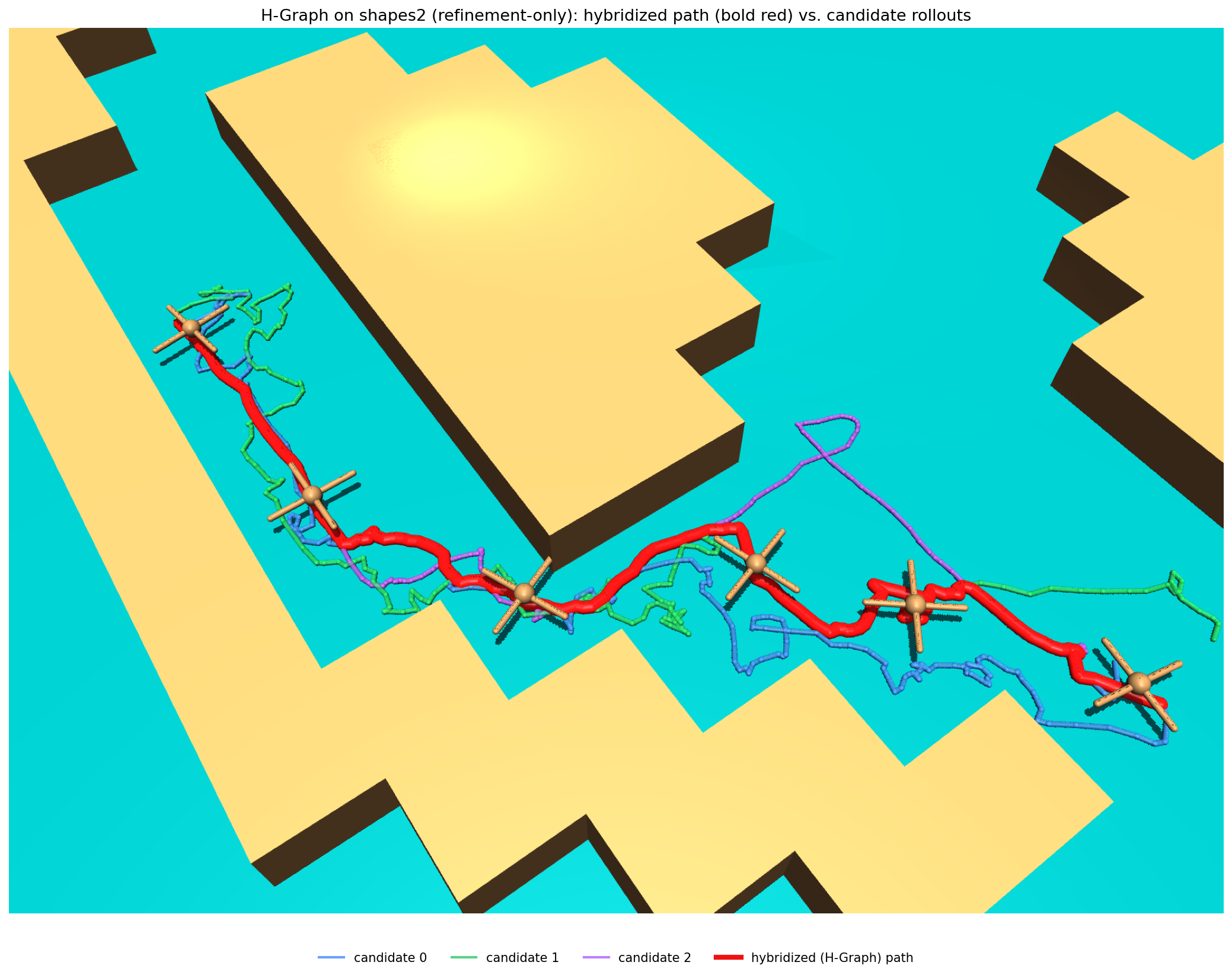}
\caption{H-Graph hybridization on shapes2 (RS sweep, seed=42, DGB=0.85), cropped to the corridor between the maze's two large obstacle blocks. Candidate paths range from RRT-dominated (blue, RS=0.00) to diffusion-dominated (purple, RS=1.00), with an intermediate, winding candidate (green, RS=0.75). The H-Graph path (bold red, small ants marking sampled waypoints) fuses segments from all three into a single trajectory shorter than any individual candidate.}
\label{fig:hgraph_refinement_shapes2}
\end{figure}

The DiTree (Diffusion Tree) approach integrates diffusion-based action sampling with RRT-style planners, enabling a single trained model to plan across diverse environments without retraining. A key property of diffusion models, however, is that the random seed sets the initial noise fed into the diffusion process, so running the same model again with a different seed, or with different conditioning parameters, produces a meaningfully different output \cite{li2025reliable, clemente2025twostepsdiffusionpolicyrobotic}. DiTree currently plans with only a single run, so it discards this variability and misses potential improvements in trajectory quality. In classical motion planning, H-Graph hybridization exploits this kind of path diversity by combining multiple solution paths into a better trajectory \cite{hgraph_original}.

This work investigates how inference-time variability in a pretrained DiTree model can be exploited without any retraining to generate a diverse pool of candidate trajectories suitable for H-Graph hybridization. We study two orthogonal diversity mechanisms. The first sweeps two parameters together, the random seed and the Diffusion Goal Bias (DGB), which controls whether the model is conditioned on the final goal or on a random intermediate waypoint. The second varies the refinement strength (RS), a parameter that controls how aggressively a flow-matching model \cite{lipman2022flow} edits each RRT-generated warm-start trajectory before it is executed. Concretely, the RRT planner first produces a short-horizon action proposal via linear steering. This proposal is then partially noised and given to the diffusion model, which edits it by integrating a learned velocity field, following the SDEdit approach of editing an input by partially noising it and then denoising it with a diffusion model \cite{meng2022sdeditguidedimagesynthesis}. At low RS, the model only lightly edits the warm-start, so the resulting trajectory stays close to it, curving around obstacles the way the RRT warm-start does, at high RS, the model largely replaces the warm-start, producing a compact, goal-directed trajectory instead. H-Graph hybridization then fuses these two complementary trajectories into a single, shorter one.

Our main contribution is twofold: exploiting inference-time parameter variation (random seed, Diffusion Goal Bias, and RRT warm-start refinement strength) as a training-free source of trajectory diversity, and adapting H-Graph hybridization to fuse this diversity into a single, higher-quality trajectory. This pipeline reliably outperforms both the candidate pool and the successful single-run baseline.

\section{Background and Related Work}

\subsection{Diffusion-Guided RRT (DiTree)}

DiTree \cite{hassidof2025trainonceplananywherekinodynamicmotion} leverages diffusion policies (DPs) \cite{chi2023diffusion} as informed samplers to guide kinodynamic sampling-based planners (SBPs) \cite{elbanhawi2014sampling, karaman2011sampling}, planners that, unlike purely geometric SBPs, follow the system's motion constraints (e.g., bounded velocity or turning radius) by propagating sampled control inputs through the robot's dynamic model. Concretely, a kinodynamic SBP initializes a tree from the start state $x_{start}$, then iteratively selects a tree node $x_{near}$, samples a control input, and applies \emph{forward propagation}, integrating the dynamic model to produce a collision-free sub-trajectory $\pi$ from $x_{near}$ to a new state $x_{new}$, which is then added to the tree. In Kinodynamic-RRT \cite{lavalle1998rapidly} specifically, $x_{near}$ is the node closest to a randomly sampled state $x_{rand} \in \mathcal{X}$, and the control is drawn by uniform random sampling. DiTree's core innovation is to replace this uniform control sampling with a diffusion model that acts as an informed sampler: conditioned on $\mathcal{X}_{obs}^{near}$ (the obstacles within a fixed radius of the selected state) and a target state, it directly generates candidate control sequences that steer the tree's growth toward the target state, rather than sampling controls at random. 
The diffusion model generates sequences of $N$ actions $u_{1:N}$ conditioned on the selected tree node $x_{near}$, a target state $x_{target}$, and local observations $\mathcal{X}_{obs}^{near}$ \cite{hassidof2025trainonceplananywherekinodynamicmotion}. During the RRT expansion phase, the \textbf{Diffusion Goal Bias (DGB)} modulates the conditioning: it replaces the final goal with a randomly sampled intermediate state $x_{rand}$ to improve search coverage and avoid local minima \cite{hassidof2025trainonceplananywherekinodynamicmotion}.

While this ensures collision avoidance and feasibility under the robot's dynamics, DiTree typically produces only a single solution path, leaving the potential for further trajectory optimization through path diversity unexplored \cite{hassidof2025trainonceplananywherekinodynamicmotion}.

DiTree's warm-start-refinement stage (Section~\ref{sec:editing}) builds on SDEdit, a prior editing technique \cite{meng2022sdeditguidedimagesynthesis}, which edits an input by partially noising it and then denoising it with a learned model, originally proposed for guided image synthesis. Within diffusion policies specifically, prior work \cite{park2025demodiffusion} applies this same paradigm to robot action trajectories: rather than sampling the trajectory from scratch, it partially noises a kinematically-retargeted human demonstration and denoises it with a pretrained diffusion policy, substantially outperforming both the raw retargeted trajectory and the policy sampled without a prior. The same mechanism has also been used for cross-domain trajectory adaptation: other work \cite{niu2024xted} edits state-action trajectories by noising and denoising them with a pretrained diffusion model to align a source-domain trajectory with target-domain dynamics. Another approach \cite{kang2026warmprior} replaces the Gaussian initialization of a flow-matching policy with a temporal prior built from recent action history, yielding straighter, faster generation paths. We adopt the same partial-noising-and-denoising mechanism, but apply it to a geometric RRT warm-start rather than a human demonstration or action history.

\subsection{H-Graph Hybridization}

H-Graph hybridization \cite{hgraph_original} is a classical post-processing technique used to improve trajectory quality by combining the strengths of multiple distinct solution paths. The algorithm constructs a ``hybrid graph'' from several input trajectories, identifies ``bridge'' connections where paths pass through the same homotopy class (``two paths are said to be homotopy equivalent if one path can be continuously deformed into the other, without introducing any collisions along the way'' \cite{hgraph_original}) or neighborhood, and extracts an optimal path through the resulting structure.

The process follows three fundamental stages:
\begin{enumerate}
    \item \textbf{Graph Construction}: A unified graph is built from all vertices and edges in the input trajectories.
    \item \textbf{Bridge Detection}: Potential transitions between paths are identified wherever their states fall within a predefined neighborhood distance $\delta$.
    \item \textbf{Path Extraction}: A local planner checks each bridge for feasibility, then a shortest-path algorithm (e.g., Dijkstra) finds the globally optimal path through the hybridized graph.
\end{enumerate}

Our work adapts the H-Graph framework to DiTree by leveraging the diversity induced by varying inference-time parameters, specifically random seeds, the Diffusion Goal Bias (DGB) factor, and RRT warm-start refinement strength, to generate high-quality candidate paths for hybridization. The key advantage lies in controlling the exploration-exploitation trade-off between two complementary planning behaviors. The diffusion model, trained on successful demonstrations, is inherently exploitation-biased: it steers trajectories toward the smooth, goal-directed motions it learned from those demonstrations. The underlying RRT sampler, in contrast, is exploration-biased: its random sampling probes obstacle-proximate regions the learned model would otherwise skip. The refinement strength $\lambda$ sets how much each component shapes a given edge: the higher $\lambda$ is, the more the diffusion model dominates, smoothing the trajectory and pushing it toward exploitation. Sweeping $\lambda$ therefore yields paths that differ systematically along the exploration-exploitation spectrum, and it is this spread that H-Graph exploits to fuse trajectory segments into a single, higher-quality path.

\section{Methodology}

\subsection{Diverse Path Generation}

To generate diverse solution paths for H-Graph hybridization, we exploit three sources of variability in the diffusion-guided RRT planner:

\textbf{Random Seed Variation:} The diffusion sampling process involves random noise generation at each denoising step. By varying the random seed across multiple planning runs, we obtain different action sequences even with identical initial conditions, leading to paths that explore different regions of the configuration space.

\textbf{Diffusion Goal Bias (DGB) Variation:} The DGB parameter controls the probability of conditioning the diffusion model on the final goal versus random intermediate targets. Lower DGB values promote exploration by encouraging the model to sample actions toward intermediate waypoints, while higher values focus on goal-directed behavior. By sampling paths with different DGB values, we obtain trajectories that vary in their exploration-exploitation balance, potentially discovering different homotopy classes.

\textbf{RRT Warm-Start Refinement Strength (RS) Variation:} DiTree generates each tree edge in two stages: (i) a linear steering function constructs a short-horizon RRT warm-start trajectory $u^{prior}$, and (ii) a flow-matching diffusion model \cite{lipman2022flow} \emph{edits} that warm-start by partially noising it and then integrating a learned velocity field from time $\lambda$ to $1$. The refinement strength $\lambda \in [0,1]$ controls how aggressively this diffusion editing modifies the warm-start: at $\lambda = 0$ the model leaves the RRT proposal unchanged, while at $\lambda = 1$ the warm-start is discarded entirely and the trajectory is generated from pure diffusion noise. By sweeping RS across several values we obtain paths that range from lightly edited, obstacle-proximate trajectories (low $\lambda$) to heavily edited, smooth and goal-directed ones (high $\lambda$). Because these two extremes tend to cover complementary regions of the maze, they supply H-Graph with the spatial diversity needed to form bridge connections. The technical details of the two-stage generation process are given in Sections~\ref{sec:prior} and~\ref{sec:editing} below.

\subsection{Warm-Start Generation via Linear Steering}
\label{sec:prior}
At each expansion iteration, the planner selects a target state $x_{target}$ by balancing exploration and exploitation. This choice is governed by the goal-conditioning bias $\gamma \in [0,1]$, which is precisely the Diffusion Goal Bias (DGB) parameter introduced in Section~II-A and swept in our experiments (Section~IV): with probability $\gamma$ the planner exploits by steering toward the fixed goal state $x_{goal}$, and with probability $1-\gamma$ it explores by steering toward a random state $x_{rand}$ drawn from $q(\cdot)$, a uniform distribution over the environment's state-space bounds. The steering target is therefore sampled as:
\begin{equation}
x_{target} =
\begin{cases}
x_{goal}, & \text{with probability } \gamma \\
x_{rand} \sim q(\cdot), & \text{with probability } 1-\gamma
\end{cases}
\end{equation}

Once $x_{target}$ is defined, the action generator constructs a short-horizon control proposal $u_{1:H}^{prior}$, i.e., a sequence of actions over a horizon of $H$ steps. This generator utilizes a deterministic linear steering function $g(x_t, x_{target})$, which maps the current state $x_t$ and the target $x_{target}$ to a control vector pointing from the former toward the latter. To ensure sufficient exploration and tree diversity, additive Gaussian noise $\eta_t$ is injected into the action sequence:
\begin{equation}
u_t^{prior} = g(x_t, x_{target}) + \eta_t, \quad \eta_t \sim \mathcal{N}(0, \sigma_t^2 I)
\end{equation}
where $u_t^{prior}$ is the resulting prior action at step $t \in \{1, \dots, H\}$, and $\eta_t$ is zero-mean Gaussian noise with covariance $\sigma_t^2 I$ ($I$ denoting the identity matrix). The magnitude of this noise follows a decaying schedule $\sigma_t = \sigma_0(1 - t/H)$, where $\sigma_0$ is the initial noise scale. This ensures that early actions explore the local manifold while later actions converge toward the steering target. A linear proxy simulation propagates the state $x_{t+1}$ from $x_t$ and $u_t^{prior}$ internally, ensuring the generated action sequence remains approximately synchronized with the agent's kinematics.

\subsection{Diffusion-Based Action Editing}
\label{sec:editing}

The RRT-generated warm-start is not executed directly. Rather, it serves as a structural initialization for a conditional flow matching sampler, following the SDEdit paradigm of editing an input by partially noising it and then denoising it with a learned model \cite{meng2022sdeditguidedimagesynthesis}. The editing process begins by partially noising the RRT warm-start $u^{prior}$ according to a refinement strength $\lambda \in [0, 1]$, where $\lambda = 0$ corresponds to the pure RRT output and $\lambda = 1$ to a full diffusion start. The initial state for the ODE solver is defined as:
\begin{equation}
u_\lambda = (1 - \lambda)u^{RRT} + \lambda \epsilon, \quad \epsilon \sim \mathcal{N}(0, I)
\end{equation}
The refined trajectory $u_{1:H}$ is then obtained by integrating the learned velocity field $v_\theta$ from the starting time $t = \lambda$ to the target $t = 1$:
\begin{equation}
u_{1:H} = u_\lambda + \int_{\lambda}^{1} v_\theta(u_\tau, \tau \mid x_{near}, x_{target}, \mathcal{X}_{obs}^{near}) d\tau
\end{equation}
where $v_\theta$ is the conditional vector field trained to minimize the flow matching objective. This refinement process is conditioned on the current state history $x_{near}$, the local observation map $\mathcal{X}_{obs}^{near}$, and the steering target $x_{target}$.

By ``editing'' the RRT prior through this learned flow, the model produces a refined action sequence that preserves the intentionality of the RRT expansion while ensuring the trajectory is physically feasible and satisfies environmental constraints. The final executed edge is generated by rolling out these edited controls through the true system dynamics $x_{t+1} = f(x_t, u_t)$.

In summary, this hierarchical framework provides the diverse candidate pool necessary for H-Graph hybridization. By coupling the global topological exploration of RRT with the dynamics-aware refinement of Flow Matching, the planner discovers geometrically diverse paths that typically remain within the same homotopy class. The stochasticity in both the warm-start generation and the refinement strength $\lambda$ ensures a wide variety of bridgeable segments, allowing the H-Graph to fuse disparate branches into a single, globally optimized solution.

\subsection{Robot Platform}

We evaluate our approach on \textbf{AntMaze}, which uses the standard MuJoCo Ant, an 8-DOF quadruped actuated by a hip and a knee joint on each of its four legs, navigating maze obstacles. Its state includes global position and orientation plus all joint angles, and its 8-dimensional action is the joint torques that drive those joints. Since forward motion requires coordinated, balanced leg actuation, an invalid control sequence (e.g., an unbalanced gait) can make the robot tip or stall even along a geometrically clear path, making this a high-dimensional, high-DOF kinodynamic planning problem.

\subsection{Implementation Details}

We use the pretrained DiTree diffusion model for AntMaze from \cite{hassidof2025trainonceplananywherekinodynamicmotion}, and a $20\times20$ local map. At each diffusion query, the model generates a 64-step action sequence (the \emph{prediction horizon}), of which only the first 2 actions are executed on the robot before the next query (the \emph{action horizon}). The H-Graph neighborhood distance $\delta$ (the maximum separation at which two states on different candidate paths are considered for a bridge, Section~II-B) is set as $\delta = 0.03 \times s$, where $s=4.0$ is the AntMaze environment scale, so that $\delta$ defines a fixed fraction of the maze size rather than a fixed absolute distance. The local planner used to verify bridges (short connecting sub-trajectories between candidate paths, Section~II-B) is allotted 10\% of the original per-edge time budget.

\subsection{Experimental Design}
\label{sec:expdesign}

We run a factorial experiment varying two parameters: seed $\in \{42, 142, 242, 342, 442\}$ and DGB $\in \{0.15, 0.5, 0.85\}$, yielding 15 input paths per scenario. Recall from Section~III-B that DGB is the goal-conditioning bias $\gamma \in [0,1]$: DGB$=0$ means the planner always steers toward a random intermediate state $x_{rand}$ (pure exploration), while DGB$=1$ means it always steers toward the fixed goal $x_{goal}$ (pure exploitation), the swept values (0.15, 0.5, 0.85) sample low-, balanced-, and high-goal-bias regimes rather than either extreme. We compare four methods:
\begin{itemize}
    \item \textbf{Baseline}: single run, seed=42, DGB=0.85.
    \item \textbf{Seed-only}: 5 paths varying seed at DGB=0.85.
    \item \textbf{DGB-only}: 3 paths varying DGB at seed=42.
    \item \textbf{Factorial}: all 15 seed$\times$DGB combinations.
\end{itemize}

We separately evaluate refinement strength (RS) as a diversity source. Recall from Section~III-C that RS is the diffusion editing parameter $\lambda \in [0,1]$: RS$=0$ means the diffusion model leaves the RRT warm-start essentially intact, integrating the velocity field over the full $[0,1]$ interval starting from the unedited warm-start, while RS$=1$ means the warm-start is discarded entirely and the trajectory is generated from pure diffusion noise ($u_\lambda = \epsilon$). For each of the 15 AntMaze scenarios, five paths are generated by fixing seed=42 and DGB=0.85 and sweeping RS $\in \{0.0, 0.25, 0.5, 0.75, 1.0\}$, with a 500-second per-path budget (the extended budget accounts for the additional time required to construct the RRT warm-start before diffusion editing), each hybridized independently.

\section{Results and Analysis}

Both tables below report the same three reference quantities, computed over the diversity pool for that table (factorial seed$\times$DGB in Table~\ref{tab:results}, RS sweep in Table~\ref{tab:refinement_five_scenes}). \textbf{Best ind.} is the minimum length among the successful single-run candidates in that pool, i.e.\ what a user would get by keeping only the single luckiest individual run rather than hybridizing. \textbf{Pool mean} is the mean length over the same successful candidates, i.e.\ the expected quality of one random draw from the pool. \textbf{Baseline len} is different in kind: it is the length achieved by the fixed default configuration (seed=42, DGB=0.85, and RS=0.5 where applicable) when it reaches the goal, representing what a single, non-diversified planner call would typically achieve, rather than a property of the diversity pool. On scenarios where the default configuration never reaches the goal, Baseline len is undefined (`--') and no vs.\ Baseline comparison is possible.

\subsection{Seed/DGB Factorial Diversity}

We evaluate seed and DGB (goal-bias) variation as a diversity source on \textbf{AntMaze}, using the full experimental design of Section~\ref{sec:expdesign} (baseline, seed-only, DGB-only, factorial) across all 15 scenarios (Table~\ref{tab:results}).

\begin{table*}[!t]
\centering
\caption{AntMaze results under seed$\times$DGB factorial diversity, all 15 scenarios (trajectory length, lower is better).}
\label{tab:results}
\resizebox{\textwidth}{!}{%
\begin{tabular}{lccccccc}
\toprule
\textbf{Scenario} & \textbf{Best ind.} & \textbf{Pool mean} & \textbf{Baseline len} & \textbf{H-Graph} & \textbf{vs.\ Best ind.} & \textbf{vs.\ Pool mean} & \textbf{vs.\ Baseline} \\
\midrule
Rand.\ XLarge Easy & 65.3 & 76.9 & 69.4 & \textbf{60.0} & +8.0\% & +22.0\% & +13.5\% \\
Boxes & 135.6 & 145.3 & -- & \textbf{120.0} & +11.5\% & +17.4\% & -- \\
Boxes 2 & 106.1 & 131.9 & -- & \textbf{108.2} & $-$2.0\% & +18.0\% & -- \\
Boxes 3 & -- & -- & -- & \textbf{--} & -- & -- & -- \\
Narrow & 44.3 & 47.8 & 44.7 & \textbf{38.6} & +12.9\% & +19.2\% & +13.7\% \\
Shapes & -- & -- & -- & \textbf{--} & -- & -- & -- \\
Shapes 2 & 60.1 & 63.0 & 63.9 & \textbf{47.9} & +20.3\% & +24.0\% & +25.0\% \\
Rand.\ Large 2 & 62.3 & 75.8 & -- & \textbf{48.8} & +21.7\% & +35.6\% & -- \\
Rand.\ Large 3 & 72.0 & 86.1 & -- & \textbf{65.5} & +9.0\% & +23.9\% & -- \\
Rand.\ XLarge 2 & -- & -- & -- & \textbf{--} & -- & -- & -- \\
Rand.\ XLarge 3 & -- & -- & -- & \textbf{--} & -- & -- & -- \\
Rand.\ Huge Easy & 59.2 & 60.9 & -- & \textbf{53.9} & +9.0\% & +11.5\% & -- \\
Rand.\ Huge Medium & 32.8 & 34.0 & -- & \textbf{30.9} & +5.7\% & +9.1\% & -- \\
Rand.\ Huge Medium 2 & 65.6 & 78.5 & 90.6 & \textbf{61.3} & +6.6\% & +21.9\% & +32.4\% \\
Race Track & 93.2 & 94.0 & 95.9 & \textbf{89.6} & +3.8\% & +4.7\% & +6.5\% \\
\midrule
\textbf{Average} & -- & -- & -- & \textbf{--} & +9.7\% & +18.8\% & +18.2\% \\
\bottomrule
\end{tabular}}
\end{table*}

Baseline succeeds in only \textbf{25\%} of these independent attempts (11/44) across the 15 scenarios, and 4 scenarios (Boxes~3, Shapes, Rand.\ XLarge~2, Rand.\ XLarge~3) never yield two successful single-run candidates under any of the three pools, so no hybridization is possible there. Where a factorial pool exists (11/15 scenarios), H-Graph beats the best individual candidate in 10 of 11 cases, by an average of \textbf{9.7\%} (range $-$2.0\% on Boxes~2, the one case where H-Graph slightly underperforms the best individual candidate, to 21.7\% on Rand.\ Large~2), on the 5 scenarios where a plain baseline attempt also succeeds and a factorial pool exists, H-Graph improves on its mean length by an average of \textbf{18.2\%} (range 6.5\% to 32.4\%, on Rand.\ Huge Medium~2).

Looking at the raw magnitudes rather than H-Graph's gains, Baseline len is consistently at or beyond the top of its pool: in all 5 scenarios where it is defined, Baseline len exceeds Best ind., and in 3 of those 5 (Shapes~2, Rand.\ Huge Medium~2, Race Track) it exceeds Pool mean as well, meaning the fixed seed=42/DGB=0.85 configuration produces a below-average path there, not merely a non-optimal one. The two exceptions (Rand.\ XLarge Easy, Narrow) still place Baseline len between Best ind.\ and Pool mean, so the default configuration is never the standout candidate in its own pool. The gap between Best ind.\ and Pool mean also varies widely, from near-identical (Race Track: 93.2 vs.\ 94.0, Rand.\ Huge Medium: 32.8 vs.\ 34.0) to a roughly 20\% spread (Boxes~2: 106.1 vs.\ 131.9, Rand.\ Large~2: 62.3 vs.\ 75.8), indicating that the factorial sweep sometimes yields a tightly clustered set of comparable successes and sometimes a mix of one strong outlier alongside several weaker ones, the latter case is where H-Graph's improvement over Pool mean tends to be largest, since bridging can route around the weaker candidates.

As a brief ablation, we also isolate each axis of the factorial sweep. Restricting the pool to the 5 seed-only paths (DGB fixed at 0.85) yields two or more successful candidates in just 5 of 15 scenarios, where H-Graph beats the best individual by an average of \textbf{9.3\%} (range 1.8\% to 19.0\%), restricting to the 3 DGB-only paths (seed fixed at 42) does so in 6 of 15 scenarios, averaging \textbf{8.6\%} (range 2.6\% to 22.7\%). Both are comparable to the factorial's 9.7\% average gain, suggesting that crossing seed and DGB mainly helps not so much by improving fusion quality once a pool exists, but by making it more likely that a pool exists at all: a usable pool forms in 11/15 scenarios under the full factorial versus 5/15 and 6/15 under seed- or DGB-only, since 3 or 5 attempts along a single axis appear less likely to yield two independently successful paths than the full 15-path sweep.

\subsection{Refinement Strength (RS) Diversity}

We evaluate refinement strength (RS) as a diversity source on \textbf{AntMaze}, sweeping RS $\in \{0.0, 0.25, 0.5, 0.75, 1.0\}$ at fixed seed=42, DGB=0.85 across all 15 scenarios (Table~\ref{tab:refinement_five_scenes}).

\begin{table*}[!t]
\centering
\caption{AntMaze results under refinement-strength (RS) diversity, all 15 scenarios (trajectory length, lower is better).}
\label{tab:refinement_five_scenes}
\resizebox{\textwidth}{!}{%
\begin{tabular}{lccccccc}
\toprule
\textbf{Scenario} & \textbf{Best ind.} & \textbf{Pool mean} & \textbf{Baseline len} & \textbf{H-Graph} & \textbf{vs.\ Best ind.} & \textbf{vs.\ Pool mean} & \textbf{vs.\ Baseline} \\
\midrule
Rand.\ XLarge Easy & 70.5 & 76.3 & -- & \textbf{67.5} & +4.2\% & +11.5\% & -- \\
Boxes & -- & -- & -- & \textbf{--} & -- & -- & -- \\
Boxes 2 & 142.5 & 142.5 & -- & \textbf{142.5} & +0.0\% & +0.0\% & -- \\
Boxes 3 & -- & -- & -- & \textbf{--} & -- & -- & -- \\
Narrow & 43.5 & 58.4 & 54.2 & \textbf{41.3} & +5.0\% & +29.3\% & +23.7\% \\
Shapes & 193.3 & 193.3 & -- & \textbf{193.3} & +0.0\% & +0.0\% & -- \\
Shapes 2 & 57.8 & 84.1 & -- & \textbf{53.2} & +7.9\% & +36.7\% & -- \\
Rand.\ Large 2 & 69.9 & 69.9 & -- & \textbf{69.9} & +0.0\% & +0.0\% & -- \\
Rand.\ Large 3 & 82.1 & 85.0 & -- & \textbf{76.7} & +6.6\% & +9.8\% & -- \\
Rand.\ XLarge 2 & 114.4 & 121.9 & 114.2 & \textbf{98.8} & +13.6\% & +18.9\% & +13.5\% \\
Rand.\ XLarge 3 & 99.9 & 99.9 & -- & \textbf{99.9} & +0.0\% & +0.0\% & -- \\
Rand.\ Huge Easy & -- & -- & -- & \textbf{--} & -- & -- & -- \\
Rand.\ Huge Medium & 47.3 & 48.8 & 50.8 & \textbf{33.7} & +28.9\% & +30.9\% & +33.8\% \\
Rand.\ Huge Medium 2 & 70.3 & 80.9 & 69.8 & \textbf{63.7} & +9.4\% & +21.3\% & +8.8\% \\
Race Track & 95.4 & 105.6 & -- & \textbf{89.5} & +6.2\% & +15.2\% & -- \\
\midrule
\textbf{Average} & -- & -- & -- & \textbf{--} & +6.8\% & +14.5\% & +19.9\% \\
\bottomrule
\end{tabular}}
\end{table*}

Baseline succeeds in \textbf{42\%} of these attempts (8/19), 3 scenarios (Boxes, Boxes~3, Rand.\ Huge Easy) never produce two successful RS candidates, so H-Graph has no pool to bridge there. Where a pool exists (12/15 scenarios), H-Graph matches or beats the best individual RS candidate in every case (0\% on the four scenarios with only one surviving RS value, up to \textbf{28.9\%} on Rand.\ Huge Medium).

The relationship between these three columns looks different under RS diversity. Where Baseline len is defined (4 scenarios), it is not always longer than Best ind.\ as it was under seed/DGB: on Rand.\ XLarge~2 and Rand.\ Huge Medium~2 the fixed RS=0.5 configuration is actually the shortest of all sampled candidates (114.2 vs.\ Best ind.\ 114.4, and 69.8 vs.\ Best ind.\ 70.3, respectively), and on both it comfortably beats Pool mean, only on Narrow and Rand.\ Huge Medium does it fall between Best ind.\ and Pool mean, or above Pool mean entirely (50.8 vs.\ Pool mean 48.8), mirroring the seed/DGB pattern. This suggests RS=0.5, a balanced mix of RRT warm-start and diffusion refinement, is itself a reasonably strong single choice. The most common pattern in this table, however, is Best ind.\ = Pool mean (Boxes~2, Shapes, Rand.\ Large~2, Rand.\ XLarge~3): these are the four scenarios noted above where only one RS value reaches the goal, so the pool collapses to a single candidate and H-Graph trivially reproduces it (+0.0\% in both columns).

Taken together, RS-only diversity pays off most reliably when several RS values reach the goal: the largest gains occur where at least two RS values survive with meaningfully different character, giving H-Graph genuine within-class segments to bridge.

\subsubsection{\textbf{Within-Class Fréchet Spread vs.\ H-Graph Gain}}

\begin{figure}[!t]
\centering
\includegraphics[width=\columnwidth]{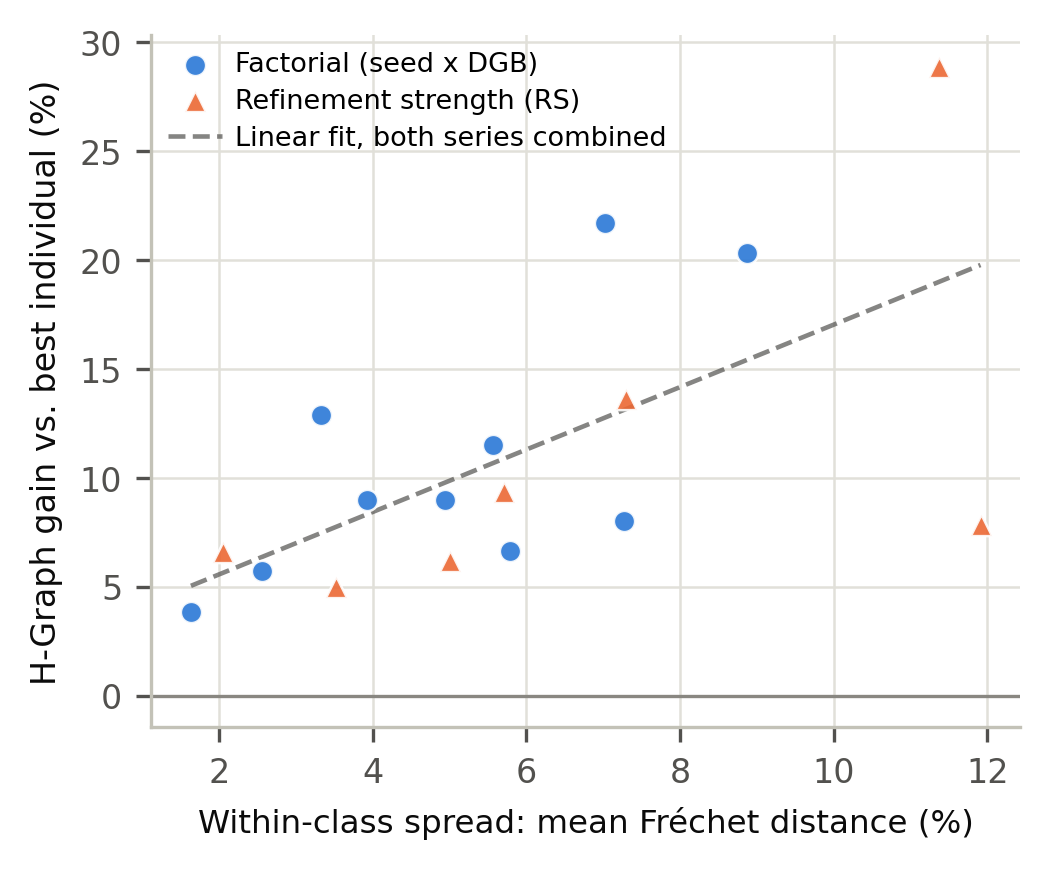}
\caption{H-Graph gain, i.e.\ \% by which H-Graph's trajectory is shorter than the best individual candidate (Tables~\ref{tab:results}, \ref{tab:refinement_five_scenes}), against each pool's within-class Fréchet spread, one point per scenario per diversity source. Dashed line: linear fit. Gain rises with spread ($r=0.63$).}
\label{fig:homotopy_within_class_frechet}
\end{figure}

To isolate the diversity that is actually hybridizable, we measure spread \emph{within} each homotopy class rather than across a whole pool. We assign each candidate to a homotopy class via its h-signature: for every free-standing maze obstacle we fix an interior point and count, with sign, how many times the candidate's trajectory crosses a fixed ray cast from that point. Candidates with identical signed-crossing counts for every obstacle belong to the same homotopy class. Within each class, we compute the discrete Fréchet distance \cite{eiter1994computing} between every pair of same-class candidates: an order-aware measure of how different two trajectories are in \emph{shape}, not just in length. We average this distance within each class, normalize by the pool's Best ind.\ length, and average across every class with at least two candidates, giving one spread value per pool (Figure~\ref{fig:homotopy_within_class_frechet}). This within-class Fréchet spread correlates with H-Graph's gain over the best individual candidate (Pearson $r=0.63$, 17 pools with at least one class of size $\geq 2$): the more same-class candidates differ in shape, the more H-Graph gains from combining them. Two pools have no usable class, every candidate there belongs to a different one, and are excluded from Figure~\ref{fig:homotopy_within_class_frechet}, one of them, Boxes~2, is also the only pool where H-Graph underperforms the best individual candidate, exactly where no same-class redundancy is left to exploit. Overall, this shows that H-Graph's gain is driven less by how many distinct classes a pool covers than by how much local shape variation exists among candidates that already share a class.

\subsubsection{\textbf{How RS Controls Trajectory Character}}

RS sets the starting time $\lambda$ of the flow-matching ODE integration, with initial condition $u_\lambda = (1{-}\lambda)\,u^{RRT} + \lambda\,\epsilon$ integrated over $[\lambda, 1]$. At low RS ($\lambda \approx 0$), the sampler inherits a nearly intact RRT warm-start and has little room to deviate, at high RS ($\lambda = 1$), it discards the warm-start and generates purely from noise.

\textbf{Low RS (RRT-dominated, $\lambda \leq 0.25$).} The RRT warm-start, built via linear steering with decaying Gaussian noise $\eta_t \sim \mathcal{N}(0,\sigma_0^2(1{-}t/H)^2 I)$, explores toward unvisited regions and so runs along obstacle boundaries. The diffusion model largely preserves this structure, producing longer trajectories that hug obstacle surfaces but cover regions a goal-directed sampler would skip.

\textbf{High RS (diffusion-dominated, $\lambda = 1$).} Without a prior, the model generates compact, direct, goal-conditioned trajectories that minimize perceived deviation. These paths are compact
in state count and take the most direct available
corridor, but they commit early to a single homotopy
class and lack exposure to alternative passages near
obstacles.

This divergence is what makes RS useful for hybridization: low- and high-RS paths cover complementary regions of the maze, creating the neighborhood overlaps H-Graph exploits as bridges.

\begin{table}[!t]
\centering
\caption{Straightness ratio (path length $\div$ straight-line distance) by scenario and RS value, lower is straighter.}
\label{tab:rs_straightness}
\resizebox{\columnwidth}{!}{%
\begin{tabular}{l c c c c c}
\toprule
Scenario & RS=0.00 & RS=0.25 & RS=0.50 & RS=0.75 & RS=1.00 \\
\midrule
Rand.\ XLarge Easy & -- & 3.21 & -- & -- & \textbf{2.81} \\
Boxes 2 & -- & 2.61 & -- & -- & -- \\
Narrow & -- & 1.40 & 1.62 & 2.81 & \textbf{1.36} \\
Shapes & -- & 1.97 & -- & -- & -- \\
Shapes 2 & 2.50 & -- & -- & 2.64 & \textbf{1.63} \\
Rand.\ Large 2 & -- & 1.98 & -- & -- & -- \\
Rand.\ Large 3 & -- & \textbf{2.72} & -- & -- & 2.95 \\
Rand.\ XLarge 2 & -- & 2.77 & \textbf{2.47} & -- & -- \\
Rand.\ XLarge 3 & -- & 10.27 & -- & -- & -- \\
Rand.\ Huge Medium & -- & -- & 2.34 & -- & \textbf{2.13} \\
Rand.\ Huge Medium 2 & -- & 1.91 & \textbf{1.53} & -- & 1.90 \\
Race Track & -- & 3.08 & -- & -- & \textbf{2.53} \\
\bottomrule
\end{tabular}%
}
\end{table}

Table~\ref{tab:rs_straightness} reports this ratio for every scenario with at least one successful RS candidate, using every one of its raw executed trajectories (12 of 15 scenarios, the omitted three never produce a successful RS candidate at all, matching Table~\ref{tab:refinement_five_scenes}). Eight scenarios have more than one successful RS value and so support a within-scenario comparison. In six of these eight, the straightest candidate is the highest RS value that was tested for that scenario (Rand.\ XLarge Easy, Narrow, Shapes~2, Rand.\ XLarge~2, Rand.\ Huge Medium, Race Track), directly supporting the RRT-to-diffusion-dominated mechanism argued above. The remaining two are exceptions: Rand.\ Large~3 goes the other way (straightest at RS=0.25, least straight at RS=1.00), and Rand.\ Huge Medium~2 dips to its straightest at the intermediate RS=0.50 rather than at RS=1.00.

\subsubsection{\textbf{Visual Evidence}}

\begin{figure}[!t]
\centering
\includegraphics[width=\columnwidth]{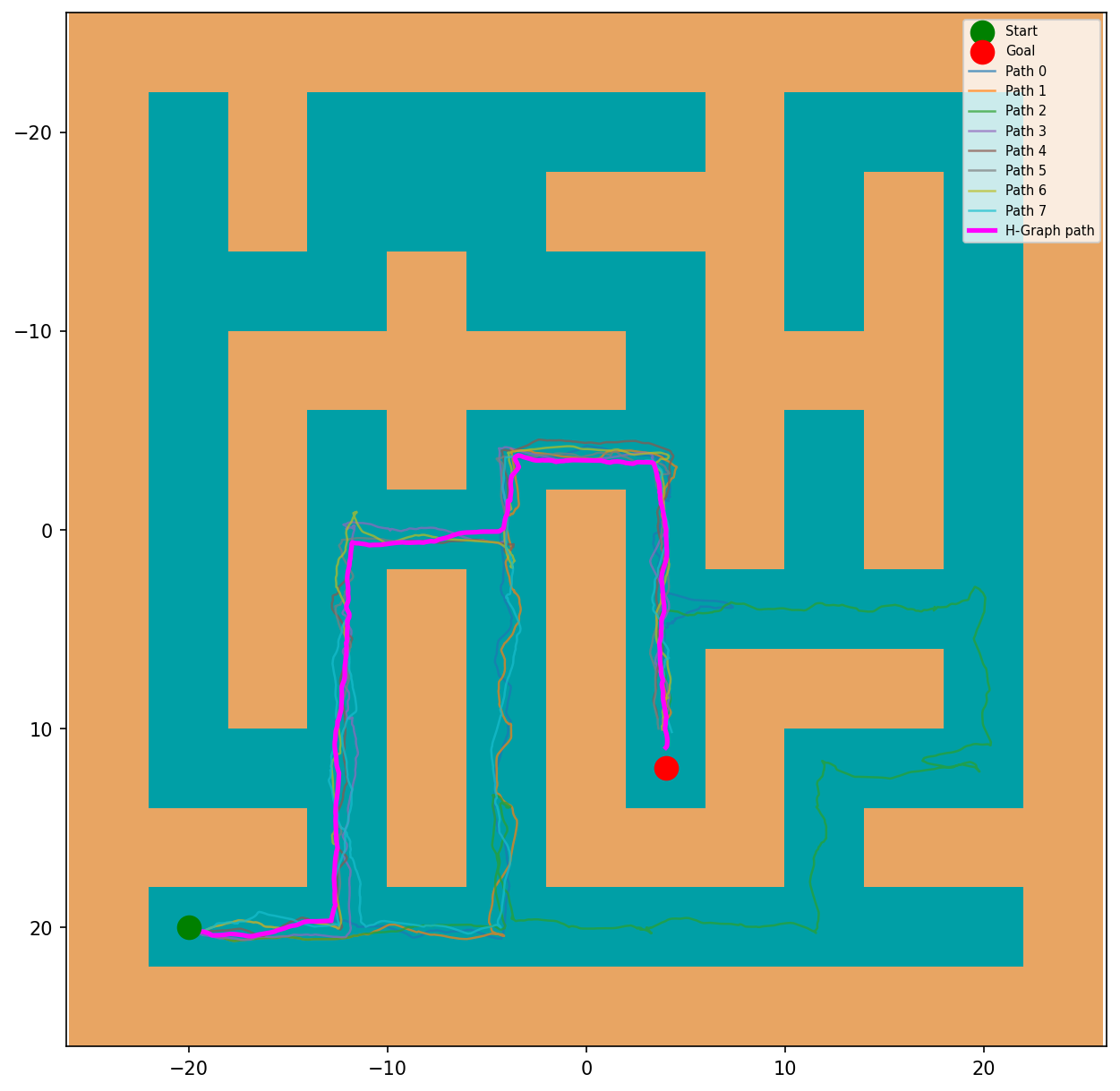}
\caption{H-Graph hybridization on Rand.\ XLarge Easy (seed$\times$DGB sweep). Four of the eight successful candidates, including the shortest (length 65.27), climb the left-hand passage near the start (green) before merging into the winding corridor toward the goal (red), three more climb the adjacent right-hand passage directly from the start baseline instead (lengths 67.40 and 77.45, plus 94.97 for the one that first backtracks out of the left-hand passage), the remaining candidate (length 106.18) also takes the right-hand passage but detours through the open area right of the corridor before rejoining it near the goal. The dense overlap downstream of the two passages yields 23{,}136 candidate bridges. The H-Graph path (magenta, 720 states, length~60.03) threads the tightest sub-segments across the pool, beating even the best individual candidate (65.3) by 8.0\%.}
\label{fig:hgraph_factorial_xlarge_easy}
\end{figure}

\begin{figure}[!t]
\centering
\includegraphics[width=\columnwidth]{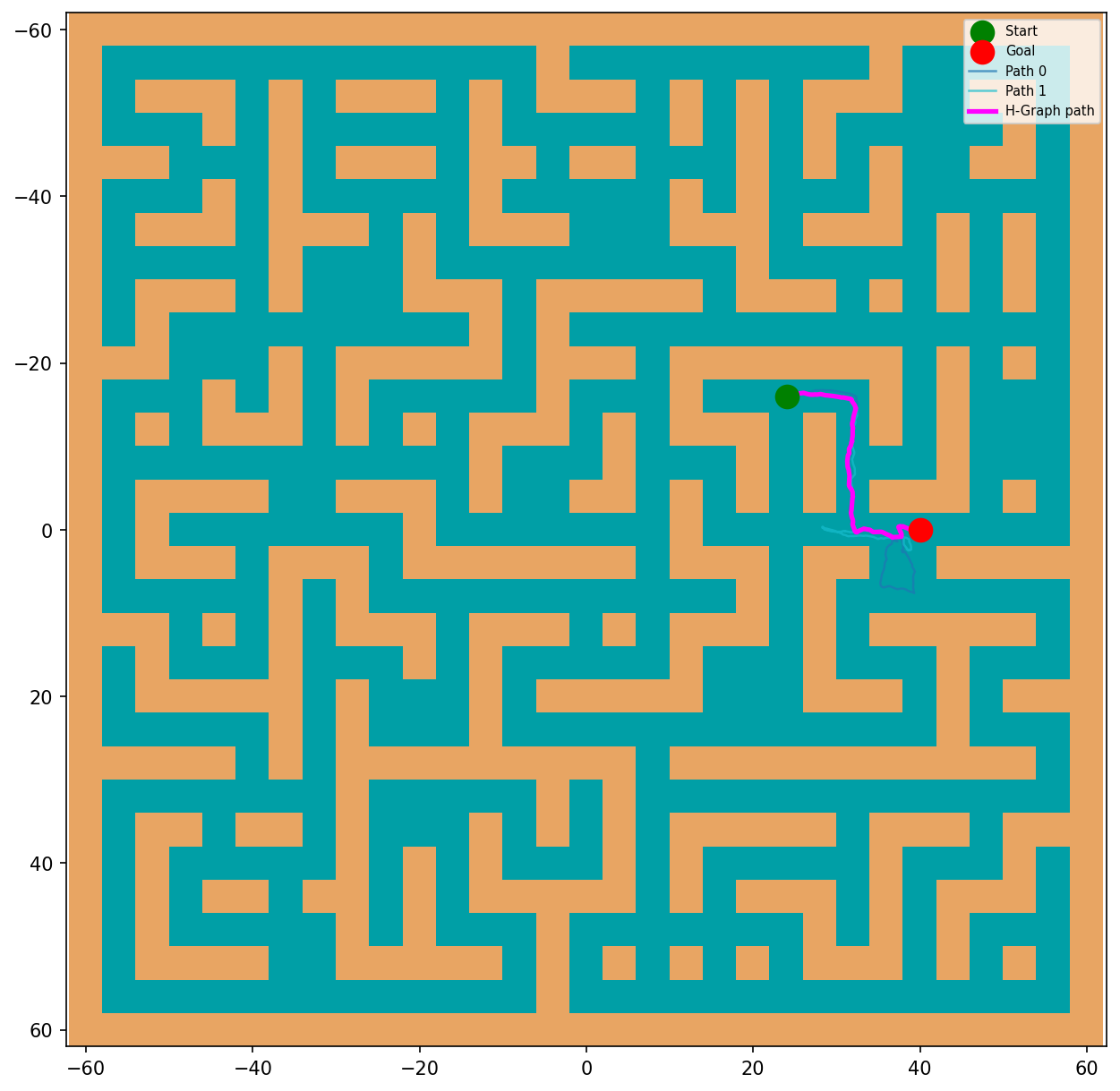}
\caption{H-Graph hybridization on Rand.\ Huge Medium (RS sweep, seed=42, DGB=0.85, only RS=0.50 and RS=1.00 reach the goal). Path~0 (RS=0.50, 829 states, length~50.34) and Path~1 (RS=1.00, 830 states, length~47.34) trace the same long vertical corridor almost identically, but each makes its own unnecessary westward jog before curving back east to the goal, tracing different loops: a single wide loop below the goal for Path~0, versus two smaller loops for Path~1, one shortly before the goal and a second right at the goal. The H-Graph path (magenta, 443 states, length~33.67) connects the two candidates directly at the corridor's foot, skipping both loops entirely, along a route neither individual candidate found, beating even the best individual candidate (47.3) by 28.9\%, the largest RS-diversity gain of any scenario reported.}
\label{fig:hgraph_refinement_huge_medium}
\end{figure}

Figure~\ref{fig:hgraph_factorial_xlarge_easy} shows the seed$\times$DGB candidate pool underlying the \textbf{Rand.\ XLarge Easy} row of Table~\ref{tab:results} (Best ind.\ 65.3, Pool mean 76.9, H-Graph 60.0, +8.0\%), Figure~\ref{fig:hgraph_refinement_huge_medium} shows the exact RS candidate pool underlying the \textbf{Rand.\ Huge Medium} row of Table~\ref{tab:refinement_five_scenes} (Best ind.\ 47.3, Pool mean 48.8, H-Graph 33.7, +28.9\%), the largest gain over the best individual candidate of any scenario in that table.

Figure~\ref{fig:hgraph_factorial_xlarge_easy} illustrates the seed$\times$DGB mechanism on \textbf{Rand.\ XLarge Easy}. Unlike the RS sweep, where low- and high-RS candidates diverge systematically in character, the eight successful seed$\times$DGB candidates split by which of two parallel vertical passages near the start they take before merging into the winding corridor toward the goal. Four candidates, including the shortest individual candidate (1129 states, length~65.27, matching the Best ind.\ entry in Table~\ref{tab:results}), climb the left-hand passage. Three more candidates instead climb the adjacent right-hand passage directly from the start baseline, skipping the left-hand passage entirely, two of these three form a distinct but comparably short homotopy class (lengths 67.40 and 77.45). The third first wanders partway into the left-hand passage, hesitates, and backtracks to the baseline before committing to the right-hand passage, incurring a large detour cost (1540 states, length~94.97). The remaining candidate also climbs the right-hand passage but then traces a long, unnecessary loop through the open area right of the corridor before rejoining it near the goal (1787 states, length~106.18, the longest of the pool). Because the two passage choices still overlap heavily downstream of the split, the dense overlap along the shared portions of the corridor yields 23{,}136 candidate bridges from just 8 input paths. The H-Graph path (magenta, 720 states, length~60.03) threads the tightest sub-segments out of this dense, mostly redundant bundle while implicitly discarding the two most wasteful candidates entirely, beating even the best individual candidate by 8.0\% and the pool mean by 22.0\%, matching the Rand.\ XLarge Easy row of Table~\ref{tab:results} exactly.

The \textbf{Rand.\ Huge Medium} scenario (Figure~\ref{fig:hgraph_refinement_huge_medium}) connects the start to the goal via a single long vertical corridor that bottoms out well west of the goal, only RS=0.50 and RS=1.00 reach the goal within budget, leaving two candidates for hybridization. Both individual candidates trace this vertical corridor almost identically, differing only by small jitter, but after reaching the bottom of the corridor each makes its own unnecessary westward jog before curving back east to the goal, tracing different loops: Path~0 (RS=0.50, 829 states, length~50.34), staying closer to the more exploratory RRT warm-start at its lower RS, takes a single wide loop below the goal before arriving, while Path~1 (RS=1.00, 830 states, length~47.34, the best individual candidate) makes two smaller loops, one shortly before the goal and a second right at the goal, before finally arriving. Because the two candidates otherwise overlap almost everywhere along the shared corridor and only diverge into their loops within that same area near the corridor's foot, this pair yields 406 candidate bridges from just 2 input paths, including bridges spanning that shared area directly to states near the goal. The H-Graph path (magenta, 443 states, length~33.67) follows the shared corridor but connects the two candidates directly at the corridor's foot, skipping both loops entirely, along a route neither individual RS candidate found: a 28.9\% reduction over the best individual candidate (Path~1, RS=1.00, length~47.34), a 30.9\% reduction over the pool mean (48.8), and a 33.8\% reduction over the baseline (50.8), exactly matching the Rand.\ Huge Medium row of Table~\ref{tab:refinement_five_scenes}. This is the largest RS-diversity gain of any scenario reported.

\section{Conclusion}

This paper demonstrated that inference-time parameter variation is a practical, training-free strategy for improving trajectory quality in diffusion-guided kinodynamic planning. By treating a fixed, pretrained DiTree model as a stochastic trajectory generator rather than a deterministic solver, we unlocked two complementary diversity sources, seed/DGB variation and diffusion editing strength (RS) and showed that H-Graph hybridization can systematically exploit the resulting path diversity.

\bibliographystyle{IEEEtran}
\bibliography{references}

\end{document}